\documentclass{naturep}

\usepackage{amssymb}
\usepackage{amsmath}
\usepackage{graphicx}
\usepackage{algorithmicx}
\usepackage{algorithm}

\usepackage[noend]{algpseudocode}
\usepackage{bbm}
\usepackage{lineno}
\usepackage[margin=0.6in]{geometry}
\usepackage{ragged2e}
\usepackage{setspace}
\usepackage{longtable}
\usepackage{hyperref}
\usepackage{anyfontsize}
\usepackage{setspace}
\usepackage{float}
\usepackage{booktabs}
\usepackage{multirow}
\usepackage{xcolor}
\usepackage{blindtext}
\usepackage{tabularx}
\usepackage{caption}
\usepackage{subcaption}

\usepackage{tabulary}
\newcolumntype{P}[1]{>{\RaggedRight\arraybackslash}p{#1}}

\makeatletter
\let\saved@includegraphics\includegraphics
\AtBeginDocument{\let\includegraphics\saved@includegraphics}

\makeatother

\newcommand{\rjc}[1]{{\textcolor[rgb]{0,0,0}{#1}}}

\title{\begin{flushleft}{\begin{spacing}{1}Algorithm Fairness in AI for Medicine and Healthcare\end{spacing}}\end{flushleft}}

\begin{document}

\maketitle
\begin{spacing}{1.8}
\vspace{-15mm}
\noindent Richard J. Chen$^{1,2,3,4}$, Tiffany Y. Chen$^{1,3}$, Jana Lipkova$^{1,2,3}$, Judy J. Wang$^{1,5}$, Drew F.K. Williamson $^{1,3}$, Ming Y. Lu$^{1,3,4,6}$, Sharifa Sahai$^{1,2,3,7}$, and Faisal Mahmood$^{*1,3,4,8,9}$
\begin{affiliations}
 \item Department of Pathology, Brigham and Women's Hospital, Harvard Medical School, Boston, MA
 \item Department of Biomedical Informatics, Harvard Medical School, Boston, MA
 \item Cancer Program, Broad Institute of Harvard and MIT, Cambridge, MA 
 \item Cancer Data Science Program, Dana-Farber Cancer Institute, Boston, MA
 \item Boston University School of Medicine, Boston, MA
 \item Electrical Engineering and Computer Science, Massachusetts Institute of Technology (MIT), Cambridge, MA
 \item Department of Systems Biology, Harvard Medical School, Boston, MA
 \item Department of Pathology, Massachusetts General Hospital (MGH), Harvard Medical School, Boston, MA
 \item Harvard Data Science Initiative, Harvard University, Cambridge, MA
 
\end{affiliations}
 
 
\end{spacing}
\begin{spacing}{1.4}
\noindent\textbf{*Correspondence:}\\ 
Faisal Mahmood \\
60 Fenwood Road, Hale Building for Transformative Medicine\\
Brigham and Women's Hospital, Harvard Medical School\\
Boston, MA 02445\\
faisalmahmood@bwh.harvard.edu
\end{spacing}

\newpage

\noindent\textbf{\Large{Abstract}}
\begin{spacing}{1.5} 
\textbf{In the current development and deployment of many artificial intelligence (AI) systems in healthcare, algorithm fairness is a challenging problem in delivering equitable care. Recent evaluation of AI models stratified across race sub-populations have revealed inequalities in how patients are diagnosed, given treatments, and billed for healthcare costs. In this perspective article, we summarize the intersectional field of fairness in machine learning through the context of current issues in healthcare, outline how algorithmic biases (\textit{e.g. -} image acquisition, genetic variation, intra-observer labeling variability) arise in current clinical workflows and their resulting healthcare disparities. Lastly, we also review emerging technology for mitigating bias via federated learning, disentanglement, and model explainability, and their role in AI-SaMD development.}
\end{spacing}

\noindent\textbf{\Large{Introduction}}

\begin{spacing}{1.5} 
Healthcare disparities continue to exist in medicine as a reflection of historical and current socioeconomic inequities, as well as group biases from the perpetuation of cultural stereotypes\cite{adler2016addressing, phelan2015racism, yehia2020association, lopez2021racial}. Though traditionally viewed through the lens of race and ethnicity, healthcare disparities encompass a wide range of dimensions, including, but not limited to: socioeconomic status, insurance status, education status, language, age, gender, sexual identity/orientation, and body mass index (BMI)\cite{bonvicini2017lgbt, yamada2015access, moy2005compiling, balsa2003clinical}. These disparities often encompass all 5 domains of the social determinants of health as defined by the US Department of Health and Human Services (economic stability, education access and quality, healthcare access and quality, neighborhood and built environment, and social and community context), which were first established to begin disentangling causal factors for worsening health outcomes and mistrust in the healthcare system\cite{marmot2005social, adler2016addressing, maness2021social, seligman2010food}. Historically, healthcare disparities began to become more widely recognized in the early 2000s with Surgeon General's reports documenting the disparities in tobacco use and access to mental health care by different racial and ethnic groups \cite{thun2000smoking}. Another example can be seen in maternal morbidity, in which trends in pregnancy-related mortality in the US when stratified by race/ethnicity showed significantly higher pregnancy-related deaths amongst non-Hispanic Black women due to disparate healthcare access and poor economic stability\cite{ronsmans2006maternal, macdorman2017trends}. A more recent example is seen in the calculation of estimated glomerular filtration rate (eGFR) as a clinical biomarker and diagnostic tool for chronic kidney disease (CKD)\cite{matsuo2010current, delanaye2013applicability, diao2021race}. Including race in this equation has led to an overestimation of kidney function in Black patients and directly affect their standard of care\cite{van2021removing, diao2021removing, vyas2020hidden}. Understanding the sources of these disparities would guide public policy on not only developing new clinical criteria for early detection of under-served patients, but also regulating the current development of machine learning algorithms trained with biased data constructed from historical inequities.

With the recent proliferation of AI algorithms in clinical deployment, there is a large ethical concern regarding the disparate impact these models will have at deployment time, especially on ethnic minority subpopulations and other underrepresented communities\cite{buolamwini2018gender, obermeyer2019dissecting, pierson2021algorithmic, hooker2021moving, mccradden2020ethical, mhasawade2021machine, currie2020ethical, chen2020ethical}. Recent audit studies have shown that AI algorithms may discover spurious causal structure in the data that correlate with protected identity status, leading to: 1) privacy leakage of race / ethnicity, and 2) misdiagnoses in using ethnicity as a shortcut for predicting outcome\cite{howard2021impact, seyyed2020chexclusion, pierson2021algorithmic}. For instance, on pathology images, recent work has shown that H\&E stain intensity is able to predict ethnicity on The Cancer Genome Atlas (TCGA) due to site-specific stain protocols and region-specific demography data. On radiology images, Convolutional Neural Networks (CNNs) will routinely under-diagnosis Hispanic patients at a disproportionate rate due to insurance type and potential lack of access to healthcare\cite{seyyed2020chexclusion}. Despite these large disparities in performance, there is a lack of public policy on regulating AI algorithms in U.S. Federal Food and Drug Administration (FDA) approval pathways to train and evaluate on diverse protected subgroups such as self-reported race and ethnicity. With 70 algorithms to date having received approval from the FDA as AI-based software as a medical devices (AI-SaMDs), AI is paced to automate many clinical paradigms that involve subjective human interpretation for disease diagnosis, prognosis, and treatment over the next decade, and if left unchecked will amplify many existing healthcare inequalities that already impact under-served subpopulations\cite{us2019proposed, gaube2021ai, us2021artificial}.

In this Perspective, we build on previous work by examining current challenges in algorithm fairness from the perspective of medical dataset shift in AI-SaMD deployment. Though health disparities and algorithm fairness are both well-reviewed areas, much of the health fairness discussion has been focused on the choice of data inductive biases such as race-specific covariates in simple risk calculators, and overlooks the broader challenges in developing fault-tolerant AI algorithms in medicine\cite{van2021removing, diao2021removing}. Moreover, while current statistical frameworks for fairness aim at learning invariance to protected class identity, such models ignore causal structure between latent biological factors such as ancestry and their associated diseases across ethnic subpopulations\cite{mccradden2020ethical}. Our Perspective begins by first providing a succinct background of fair machine learning and current fairness criteria, followed by a discussion on how genetic variation, differences in image acquisition techniques, and evolving population shifts will become obstacles in evaluation of AI-SaMDs at deployment time. \rjc{Lastly, we highlight emerging technologies and advances (federated learning, disentanglement, model interpretability) that can be used for mitigating bias and improving fairness evaluation in  healthcare, as well as their impact in AI-SaMD development. A glossary of terms of given in \textbf{Box 1}.}

\begin{center}
\begin{spacing}{1} 
\fbox{\begin{minipage}{42em}
\textbf{Box 1. Glossary of Terms.}\\

\noindent\rjc{\textbf{Health Disparities:} Group-level inequalities as a result of socioeconomic factors and social determinants of health, such as: insurance status, education level, average income in Zipcode,  language, age, gender, sexual identity / orientation, BMI.}\\

\noindent\rjc{\textbf{(Self-Reported) Race:} A recently-evolved human construct in categorizing human populations, overloading taxonomies such as ancestry, ethnicity, and nationality.}\\

\noindent\rjc{\textbf{Protected / Sensitive Attributes:} Patient-level metadata that we would like our algorithm to be non-discriminatory against in predicting outcomes.}\\

\rjc{\noindent\textbf{Protected Subgroup:} A group of patients under the same category of a protected attribute type.}\\

\noindent\rjc{\textbf{Disparate Treatment:} Intentional discrimination of protected subgroups. Disparate treatment can result from machine learning algorithms that include sensitive attribute information as direct input, or have confounding features that explain the protected attribute.}\\

\noindent\rjc{\textbf{Disparate Impact:} Unintentional discrimination as a result of disproportionate impact on protected subgroups.}\\

\noindent\rjc{\textbf{Fairness:} A recently developed statistical field that formalizes minimizing disparate treatment and impact via quantifiable fairness criteria. These fairness criteria are quantified via evaluating differences in performance metrics (\textit{e.g.} - accuracy, TPR, FPR, risk measures) across protected subgroups, as defined in \textbf{Box 2}. Larger differences would attribute larger disparate impact.}  \\

\noindent\rjc{\textbf{AI-SaMD:} Artificial Intelligence-based Software as a Medical Devices, a categorization of medical devices undergoing regulation by the U.S. Food and Drug Administration (FDA).}\\

\noindent\rjc{\textbf{Model Auditing:} Post-hoc quantitative evaluation which assesses violations of fairness criteria, often coupled with explainability techniques which attribute influential features in fairness}.\\

\noindent\rjc{\textbf{Dataset Shift:} Mismatch in train (source) and test (target) dataset distributions.}\\

\noindent\rjc{\textbf{Domain Adaptation:} Techniques that correct for dataset shift in the source and target distribution. Domain adaptation methods typically match the input spaces of the source and target distribution via techniques such as importance weighting, or in the feature space via adversarial learning.}\\

\noindent\rjc{\textbf{Federated Learning:} A form of \rjc{privacy-preserving} distributed learning that trains neural networks on local clients and send update weight parameters to a centralized server, without sharing data.}\\

\noindent\rjc{\textbf{Fair Representation Learning:} A subfield in deep learning that aims at learning intermediate feature representations invariant to protected attributes, typically self-supervised and using adversarial learning.}\\

\noindent\rjc{\textbf{Disentanglement:} A property of intermediate feature representations from deep neural networks, in which individual features control independent sources of variation in the data.}

\end{minipage}}
\end{spacing}
\end{center}

\clearpage

\noindent\textbf{\Large{\rjc{Fairness and Machine Learning}}}

\noindent\textbf{\large{\rjc{Definition and Criteria}}}

\rjc{The axiomatization of fairness is a collective societal problem that has existed beyond machine learning and healthcare.} In legal history, fairness first emerged as a research problem in developing non-discrimination laws such as the 1964 United States (U.S.) Civil Rights Act, which has made it illegal to discriminate based on legally-protected protected classes (e.g. - race, color, sex, national origin) in Federal programs (Title VI) and employment (Title VII). \rjc{In developing and evaluating non-discrimination in algorithms, the issue of fairness is also a longstanding discussion in many other domains, with algorithmic fairness and AI transparency in particular recently becoming central issues in the distribution of justice in governance\cite{feller2016computer, dressel2018accuracy}, diversity hiring in recruitment, the development of moral machines in autonomous vehicles\cite{awad2018moral}, and now the increasing deployment of AI algorithms in software medical devices and healthcare systems\cite{char2018implementing}.}

\rjc{In the machine learning and statistics community, current frameworks for understanding fairness aim at learning neutral models that are: 1) invariant to protected class identities when predicting outcomes (disparate treatment), and 2) have non-discriminatory impact on protected subgroups with equalized outcomes (disparate impact)\cite{corbett2018measure}. Formally, for a given sample with features $X$ with target label $Y$, let $A$ be a protected attribute that denotes a sensitive characteristic about the population of sample $X$ that we want our model $P(Y|X)$ to be non-discriminatory against during evaluation in predicting $Y$. To mitigate disparate treatment in algorithms, the most intuitive but naive approach would be "fairness through unawareness", in which knowledge of $A$ is denied from the model $P(Y|X)$ in making a prediction $R$.}

\rjc{However, denying protected attribute information has been shown to be insufficient in non-discrimination and satisfying fairness guarantees for many applications, as other input features may be unknown confounders that correlate with protected group membership\cite{calders2009building, chen2019fairness}. As a canonical example, the 1998 COMPAS algorithm is a risk tool that excludes race as a covariate in predicting criminal recidivism, and was contended to be fair in mitigating disparate treatment as predictions were made independent of race. However, despite not using race as a covariate, a recent retrospective study found that of defendants who did not reoffend, black defendants were approximately twice-as-likely to be assigned medium-to-high risk scores (44\%) than that of white defendants (22\%) by COMPAS\cite{feller2016computer}.} This example illustrates how differing notions of disparate impact are in conflict which one another, which has since motivated the ongoing development of formal definitions of group fairness evaluation criteria in supervised learning algorithms, as shown in \textbf{Box 2}\cite{kleinberg2016inherent, hardt2016equality}. For instance, whereas fairness via Demographic Parity was satisfied in demonstrating equal risk scores for white and black defendants who re-offended, Equalized Odds was violated due to unequal False Positive Rates (FPRs).


\noindent\textbf{\large{\rjc{Sources of Unfairness and Dataset Shift}}}


\rjc{Strategies for evaluating and satisfying group fairness have typically involved optimizing parity of fairness metrics across protected subgroups via modifying the input data, training objective, or output of the algorithm. For instance, importance weighting is a data preprocessing technique that reweights infrequent samples belonging to protected subgroups\cite{celis2019improved, kamiran2012data, krasanakis2018adaptive, jiang2020identifying} (\textbf{Figure 1}). Model constraints via regularization penalty terms are designed to remove unwanted confounders that would leak protected subgroup identity\cite{celis2019improved, zhang2018mitigating, kim2018fairness}. However, though reducing violation of fairness criteria, these techniques are insufficient in addressing systemic healthcare inequities that would bias data-generating processes. In fact, recent quantitative assessment of fairness techniques have found that optimizing parity of fairness metrics resulted in worse model performance across all subgroups in several healthcare applications, which is often described as an accuracy-fairness trade-off~\cite{zhao2019inherent, pfohl2019creating, pfohl2021empirical, pfohl2021recommendations}.}

\rjc{In order to develop domain-specific frameworks for mitigating harm in healthcare and medicine, we turn towards pinpointing root sources of unfairness and their contributions to emerging challenges in AI-SaMD deployment. Our perspective is that healthcare disparities as a result of medical AI can be better understood as dataset shift problems in which differences in population demographics and genetics, image acquisition techniques, disease prevalence, and social determinants of health would violate i.i.d. assumptions and cause disparate performance at test time~\cite{quinonero2009dataset, subbaswamy2019preventing, subbaswamy2020development, maness2021social, guo2022evaluation, bernhardt2022investigating}. Dataset shift occurs when there is a mismatch between the training and testing data distributions during algorithm development, \textit{e.g.} - $P_{\text{train}}(X) \neq P_{\text{test}}(X)$, and has important intersections with fairness as differences between train and test distributions at the subgroup-level may lead to disparate performance\cite{guo2022evaluation, ghosh2021faircanary, mishler2022fair, sagawa2019distributionally, rezaei2021robust}. In fact, fairness techniques such as importance weighting and adversarial learning were initially introduced as dataset shift techniques in mitigating covariate / domain shift, as strong assumptions such as train and test datasets being independently and identically drawn (i.i.d.) from the same distribution often do not hold at deployment time\cite{sugiyama2007covariate, buda2018systematic,ganin2015unsupervised}.}

\rjc{Group unfairness via dataset shift is especially apparent in "black-box" AI algorithms developed for structured modalities such as images and text, in which ML or clinical practitioners are unaware of domain-specific cues that would leak subgroup identity in the input\cite{bernhardt2022investigating}. For instance, in developing an AI algorithm trained on cancer pathology data from the United States and deployed on data from Turkey, domain shifts as a result of variation in H\&E staining protocols, as well as population shifts due to imbalanced ethnic minority representation, may cause the model to severely misdiagnose Turkish cancer patients. In other cases of dataset shift, hospitals may operate with different International Classification of Disease (ICD) systems, which results in label shifts in how algorithms are evaluated\cite{tedeschi1984classification, heslin2017trends}. Overall, algorithms that would be sensitive to dataset shift during deployment, may also be prone to exacerbating healthcare disparities and under-perform on fairness metrics. Thus, there are limitations in how AI models can be fairly evaluated in real-world clinical settings because of the unavailability of ground truth labels at test time. }

\vspace{10mm}

\begin{table}[h]
\footnotesize
\begin{tabular}{|P{0.1\linewidth} |P{0.15\linewidth} |P{0.07\linewidth} |c| P{0.04\linewidth} P{0.04\linewidth} P{0.04\linewidth} P{0.04\linewidth} P{0.04\linewidth} P{0.04\linewidth} P{0.04\linewidth} |c|}
\toprule
\textbf{Dataset} & \textbf{Modalities} & \textbf{Num. Patients} & \textbf{Female} & \textbf{W} & \textbf{B} & \textbf{A} & \textbf{HL}  & \textbf{PH} & \textbf{IA} & \textbf{O} & \textbf{Audit}  \\
\midrule
MSK-IMPACT\cite{cheng2015memorial} & Genomics & 10336 & 0.502 & - & - & - & - & - & - & - & N/A \\
TCGA\cite{liu2018integrated} & Pathology, MRI/CT, Genomics & 10953 & 0.485 & 0.675 & 0.079 & 0.059 & 0.003 & 0.001 & 0.002 & - & \cite{howard2021impact} \\
UK Biobank\cite{sudlow2015uk} & Genomics & 503317 & 0.544 & 0.946 & 0.016 & 0.023 & - & - & - & 0.015 & \cite{puyol2021fairness} \\
PIONEER\cite{shi2014prospective} & Genomics & 1482 & 0.434 & - & - & 1.000 & - & - & - & - & N/A \\
eMerge Network\cite{mccarty2011emerge, gottesman2013electronic} & Genomics & 20247 & - & 0.777 & 0.161 & 0.001 & - & 0.002 & 0.002 & 0.045 & \cite{li2021targeting} \\
NHANES\cite{foley2005cardiovascular} & Lab Measurements & 15560 & 0.504 & 0.339 & 0.263 & 0.105 & 0.227 & - & - & 0.065 & \cite{diao2021race,pan2021explaining} \\
Undisclosed EMR Data\cite{obermeyer2019dissecting} & EMRs, Billing Transactions & 49618 & 0.629 & 0.877 & 0.123 & - & - & - & - & - & \cite{obermeyer2019dissecting} \\
OAI\cite{nevitt2006osteoarthritis} & Limb XR & 4172 & 0.574 & 0.709 & 0.291 & - & - & - & - & - & \cite{vaughn2019racial, pierson2021algorithmic} \\
SIIM-ISIC\cite{rotemberg2021patient} & Dermoscopy & 2056 & 0.480 & - & - & - & - & - & - & & \cite{kinyanjui2019estimating, kinyanjui2020fairness} \\
NIH AREDS\cite{chew2012age} & Fundus Photography & 4203 & 0.567 & 0.977 & 0.014 & 0.080 & 0.020 & 0.012 & 0.010 & - & \cite{joshi2021ai} \\
\end{tabular}
\end{table}

\begin{table}[]
\footnotesize
\begin{tabular}{|P{0.1\linewidth} |P{0.15\linewidth} |P{0.07\linewidth} |c| P{0.04\linewidth} P{0.04\linewidth} P{0.04\linewidth} P{0.04\linewidth} P{0.04\linewidth} P{0.04\linewidth} P{0.04\linewidth} |c|}
\textbf{Dataset (Cont.)} & \textbf{Modalities} & \textbf{Num. Patients} & \textbf{Female} & \textbf{W} & \textbf{B} & \textbf{A} & \textbf{HL}  & \textbf{PH} & \textbf{IA} & \textbf{O} & \textbf{Audit}  \\
\midrule
RadFusion\cite{zhou2021radfusion} & EMRs, CT & 1,794 & 0.521 & 0.626 & - & - & - & - & - & 0.374 & \cite{zhou2021radfusion} \\
CPTAC\cite{edwards2015cptac} & Pathology, Proteomics & 2,347 & 0.395 & 0.365 & 0.032 & 0.100 & 0.023 & 0.001 & 0.004 & 0.491 & N/A \\
MIMIC\cite{johnson2016mimic, johnson2016mimic} & Chest XR, EMRs, Waveforms & 43,005 & 0.441 & 0.682 & 0.092 & 0.029 & 0.04 & 0.002 & 0.002 & & \cite{boag2018racial, meng2021mimic, panigutti2021fairlens, seyyed2020chexclusion, roosli2022peeking, pfohl2021empirical} \\
CheXpert\cite{irvin2019chexpert} & Chest XR & 64,740 & 0.410 & 0.670 & 0.060 & 0.130 & - & - & - & 0.113 & \cite{seyyed2020chexclusion, meng2021mimic} \\
NIH NLST\cite{national2011national} & Chest XR, Spiral CT & 53,456 & 0.410 & 0.908 & 0.044 & 0.020 & 0.017 & 0.004 & 0.004 & 0.020 & \cite{meng2021mimic, prosper2021association} \\
RSPECT\cite{colak2021rsna} & CT & 270 & 0.530 & 0.900 & 0.100 & - & - & - & - & - & \cite{meng2021mimic} \\
DHA\cite{gertych2007bone} & Limb XR & 691 & 0.492 & 0.520 & 0.482 & - & - & - & - & - & \cite{meng2021mimic} \\
\rjc{EMBED\cite{jeong2022emory}} & Mammography & 115,910 & 1.000 & 0.389 &  0.416 & 0.065 & 0.056 & 0.010 & - & 0.113 & N/A \\
\rjc{Optum\cite{gertych2007bone}} & EMRs, Billing Transactions & 5,802,865 & 0.561 & 0.670 & 0.075 & 0.028 & 0.075 & - & - & 0.152 & \cite{pfohl2021empirical} \\
\rjc{eICU\cite{pollard2018eicu}} & EMRs & 200,859  & 0.460 & 0.773 & 0.106 & 0.016 & 0.037 & - & 0.009 & 0.059 & \cite{Venet2020} \\
\rjc{Heritage Health\cite{el2012identification}} & EMRs & 172,731  & 0.544 & - & - & - & - & - & - & - & \cite{Madras2018, louizos2017, Raff2018, raff2018gradient, madras2018a} \\
\rjc{Pima Indians Diabetes\cite{Smith1988}} & Population Health Study  & 768  & 1.000 & - & - & - & - & - & 1.000 & - & \cite{Sharma2020, Chen2019} \\
\rjc{Warfarin\cite{nejm2009}} & Drug Relationship & 5,052  & - & 0.553 & 0.089 & 0.303 & - & - & - & 0.054 & \cite{Kallus2020} \\
\rjc{Infant Health (IDHP)\cite{ihdp1990data}} & Clinical Measures & 985  & 0.509 & 0.369* & 0.525 & - & 0.107 & - & - & - & \cite{madras2019data, yi2019data} \\
\rjc{DrugNet\cite{weeks2002data}} & Clinical Measures & 293  & 0.294 & 0.085* & 0.338 & - & 0.529 & - & - & - & N/A \\

\bottomrule
\end{tabular}
\caption*{\textbf{Table 1. \rjc{Sex and race demography of well-known biomedical dataset benchmarks used in machine learning and fairness evaluation}.} Reported demography data were obtained for all patient populations in the original dataseat, though model auditing may use only certain subsets due to missing labels and/or insufficient samples for evaluation in extremely under-represented minorities, and/or target different protected attributes such as age, income and geography. Abbreviations: W = White, B = Black, A = Asian, HL = Hispanic / Latino, PH = Pacific Islander / Native Hawaiian, IA = American Indian / Alaskan Native. O = Unknown or Other. "-" denote demographic data not made publicly-available or acquired. \rjc{"*" denotes grouping of "White" and "Unknown / Other" together.}}
\end{table}

\newpage
\clearpage

\begin{center}
\begin{spacing}{1} 
\fbox{\begin{minipage}{42em}
\textbf{Box 2. Brief background on fairness criteria.}\\
\includegraphics[width=42em]{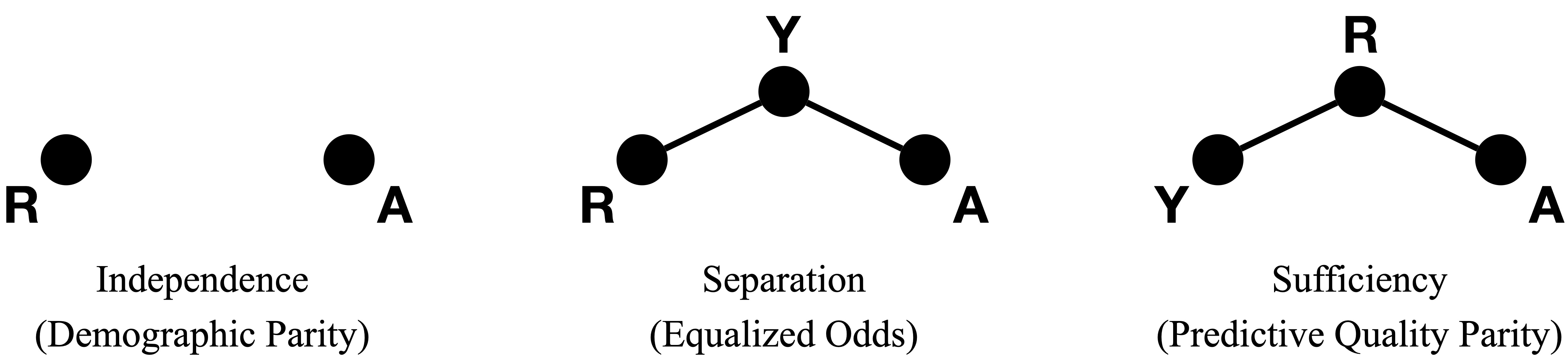}
For binary classification tasks, $Y \in \{0,1\}$ denotes a binary target label used to supervise our model, with $R \in [0,1]$ denoting the classification score $P(Y|X)$ made by our model. For clinical tasks, $Y$ can refer to objective health outcomes (\textit{e.g.} survival, response-to-treatment), or subjective clinical annotations and diagnoses (\textit{e.g.} stage, grade, or subtype of disease). In evaluating models for non-discrimination, three representative fairness metrics are used in current practice: a) Demographic Parity, b) Equalized Odds, and c) Calibration. In satisfying these fairness criterias, not all criterion can be satisfied at the same time.

\noindent\textit{a. Demographic Parity.} Demographic Parity asserts that the fraction of positive predictions made the model should be equal across protected subgroups, satisfying the \textbf{independence} criteria $R \perp A$ via the constraint:
\begin{align}
    \mathbb{P}\{R=1 \mid A=a\}=\mathbb{P}\{R=1 \mid A=b\}
\end{align}
for different subgroups $a, b$. Independence reflects the notion that decisions should be made independently of the subgroup identity. However, note that demographic parity only constrains the rate of positive predictions, and does not consider the rate at which the ground truth label may actually occurs in the subgroups. For instance, for $Y=1$ indicating kidney failure, disparate access to healthcare may self-select black patients at a relatively greater proportion than whites in the population, however, equalizing model predictions may decrease the number of black patients that are positively predicted.

\noindent\textit{b. Equalized Odds.} Equalized Odds asserts that the true positive and false positive rates (TPR, FPR) should be equalized across protected subgroup, satisfying the \textbf{separability} criteria $R \perp A | Y$ via the constraints:
\begin{align}
    \mathbb{P}\{R=1 \mid Y=1, A=a\}=\mathbb{P}\{R=1 \mid Y=1, A=b\} \\
    \mathbb{P}\{R=1 \mid Y=0, A=a\}=\mathbb{P}\{R=1 \mid Y=0, A=b\}
\end{align}
In comparison to independence,  separability states that algorithm scores should be conditionally independent of the protected attribute given the ground truth label. As a result, Equalized Odds considers that subgroups can have different distributions of $P(Y)$ and is incentivized to reduce errors uniformly across all subgroups.

\noindent\textit{c. Predictive Quality Parity.} Predictive Quality Parity asserts that the predictive positive and negative values should be equalized across subgroups, satisfying the \textbf{sufficiency} criteria $Y \perp R | A$ via the constraint:
\begin{align}
    \mathbb{P}\{Y=1 \mid R=r, A=a\}=\mathbb{P}\{Y=1 \mid R=r, A=b\}
\end{align}
For unthresholded scores, Predictive Quality Parity can be viewed as a form of "calibration by group" in which for score $r$ in the support of R, the following calibration constraint is satisfied for all subgroups in $A$:
\begin{align}
    \mathbb{P}\{Y=1 \mid R=r, A = a\}=r, \enspace \forall a \in A
\end{align}
\end{minipage}}
\end{spacing}
\end{center}

\vspace{-4mm}

\begin{figure*}
\vspace{-9mm}
\begin{center}
\includegraphics[width=0.9\textwidth]{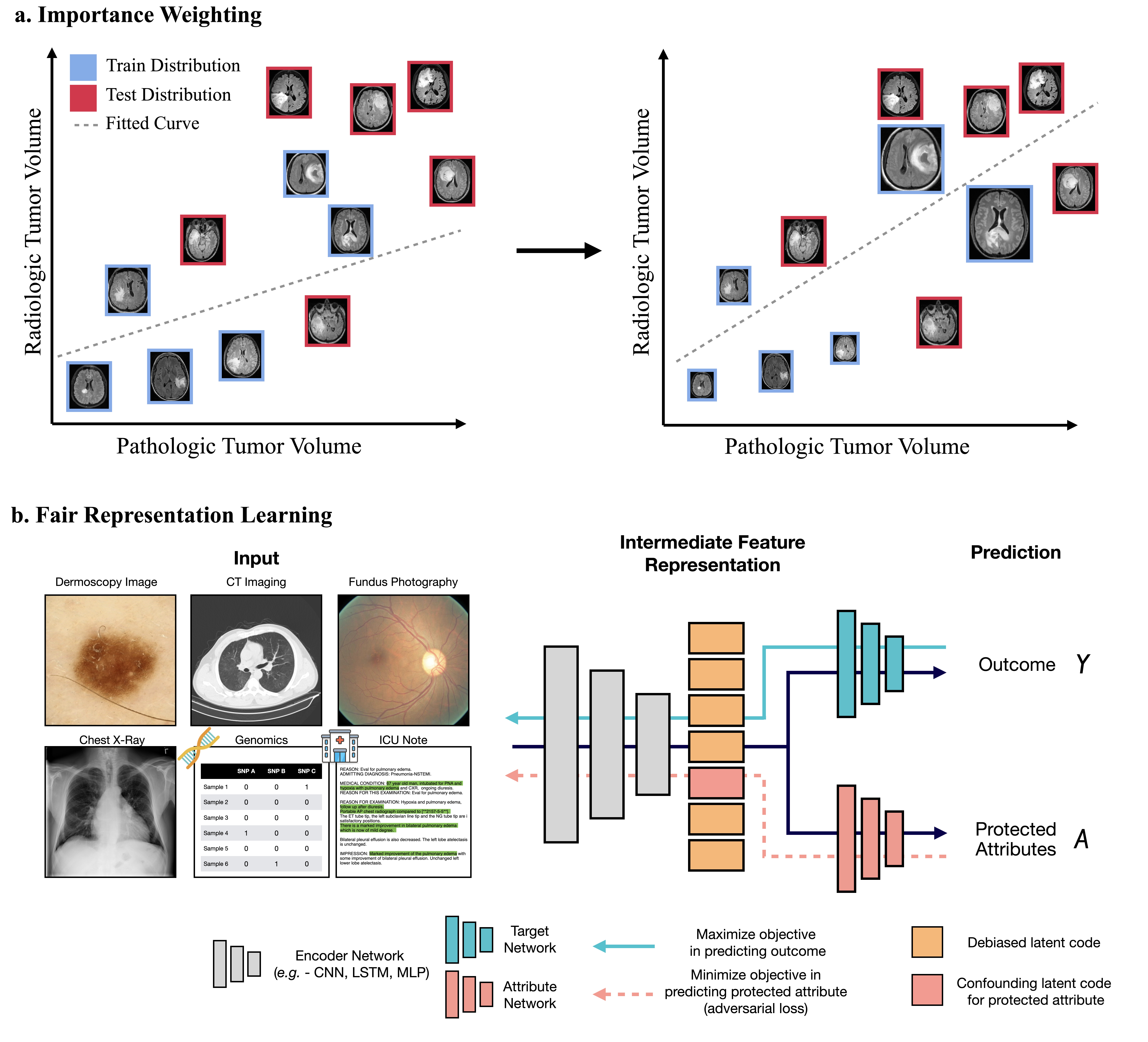}
\caption*{\textbf{Figure 1. Strategies for mitigating disparate impact.} \textbf{a.} For under-represented samples in the train and test distribution, importance weighting can be applied to reweight the infrequent samples to match the distributions. Mismatches in train and test distribution may occur in deploying an algorithm in a population with different demographies, disease prevalances, and sample selection biases that result in label prejudice. \textbf{b.} To remove protected attributes in the representation space of structured data modalities such as images and text data, deep learning algorithms can be additionally supervised with the protected attribute as a target label, in which the loss function for attribute prediction is maximized. Such strategies are additionally referred to as debiasing. In clinical machine learning tasks, modalities such as fundus photography images or chest X-ray images have been shown to include subtle biases that may leak protected attribute information such as age, gender, and self-reported race.}
\end{center}
\end{figure*}

\noindent\textbf{\Large{Emerging Challenges in AI-SaMD Deployment}}

\rjc{In this section, we examine several domain-specific challenges in current AI deployment in healthcare from the dual perspective of group fairness and dataset shift, as illustrated in the varying cases in genomics in \textbf{Figure 2} and medical imaging in \textbf{Figure 3}. A high-level overview of dataset shift with examples in medicine can be found in \textbf{Table 2}, with a formal introduction referred to other literature\cite{quinonero2009dataset, castro2020causality}.} 



\vspace{-4mm}
\noindent\textbf{\large{Missing Ethnic and Ancestral Diversity in Biomedical Datasets}}
\vspace{-4mm}

In the current development and integration of AI-based computer-aided diagnosis (CAD) systems in healthcare, the vast majority of models are trained on race-skewed datasets that over-represent individuals of European ancestry, with race-stratified evaluation largely ignored in reporting precision and recall. Moreover, much of our understanding of many diseases has been developed using biobank repositories that predominantly represent individuals with European ancestry\cite{kraft2018beyond, west2017genomics, sudlow2015uk}. Ancestry, along with other demographic data, is a crucial determination of the mutational landscape and the pathogenesis of cancer, with the prevalence of certain mutations only detectable in high-throughput sequencing of large and representative cohorts\cite{landry2018lack}. For instance, individuals with Asian ancestry are known to have a high prevalence of EGFR mutations as detected in the PIONEER cohort that enrolled 1482 Asian patients  (\textbf{Figure 2})\cite{shi2014prospective}. However, in the The Cancer Genome Atlas (TCGA), across 8,594 tumor samples from 33 cancer types, 82.0\% of all cases were from Whites, 10.1\% were from Blacks or African Americans, 7.5\% were from Asians, and 0.4\% from extremely under-reported minorities such as Hispanics, Native Americans, Native Hawaiians and other Pacific Islanders (denoted as "Other" in the TCGA) (\textbf{Table 1, Figure 2})\cite{gao2013integrative}. Due to lack of genetic diversity, many common genomic alterations such as EGFR (discovered by other high-sequencing efforts) are undetectable in the TCGA, despite being extensively used to discover molecular subtypes and redefine World Health Organization (WHO) taxonomies for cancer classification\cite{spratt2016racial, zhang2019characterization}.

Despite these disparities in representation, many AI algorithms undergoing "clinical-grade" validation are trained and evaluated on race-skewed, public biobank datasets without considering their disparate impact on minority subpopulations due to population shift. For instance, the first AI models to surpass clinical-grade performance on predicting lymph node metastases were trained and evaluated on the CAMELYON16/17 datasets sourced entirely from the Netherlands\cite{bejnordi2017diagnostic, campanella2019clinical}. However, such algorithms have yet to evaluate race-stratified performance due to a lack of large and ethnic-diverse external cohorts. For cancer types such breast cancer in which there is known genetic diversity in hormone receptor status amongst ethnic subpopulations, phenotypic manifestations of genetic diversity may leak ethnicity subgroup information in diagnostic algorithms\cite{zavala2020cancer, zhang2017racial, ooi2011disparities, henderson2012influence, gamble2021determining}. In this example, ancestry and genetic variation are latent variables that may manifest in the tissue microenvironment, which poses a challenge in the representation space and would thus, entail debiasing strategies such as adversarial learning or regularization. Many attempts in establishing histology-genomic correspondences have also been only accomplished using the TCGA and other European biobanks, which makes computational pathology and genomics a challenging domain in dataset shift and model calibration\cite{kather2019deep, kather2020pan, fu2020pan}. 


\begin{figure*}
\vspace{-9mm}
\begin{center}
\includegraphics[width=0.9\textwidth]{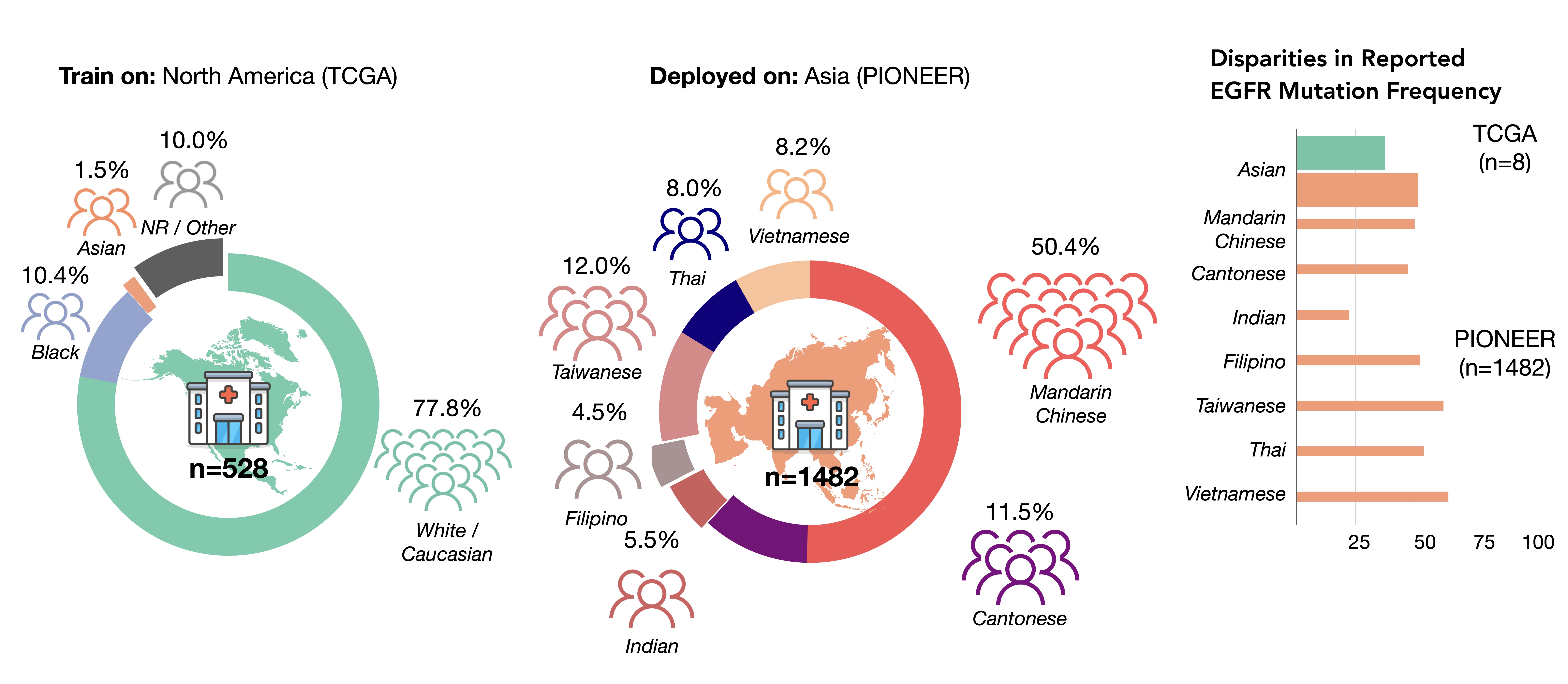}
\caption*{\textbf{Figure 2. Genetic drift as population shift.} Demography characteristics and gene mutation frequencies for EGFR of lung adenocarcinoma patients in the TCGA (green) and PIONEER (orange) cohort. Of the 528 lung adenocarcinoma patients in the TCGA, only 1.5\% ($n=8$) self-report as "Asian", versus the 1482 Asian patients enrolled in PIONEER, which includes a more fine-grained self-reported ethnicity / nationality categorization of: Mandarin Chinese, Cantonese, Taiwanese, Vietnamese, Thai, Filipino and Indian. As a result of under-representation of Asian patients in the TCGA, the mutation frequency for genes such as EGFR, which is commonly used in guiding the use of TKIs as treatment, is only 37.5\% ($n=3$). In PIONEER, overall EGFR mutation frequency for all Asian patients was found to be 51.4\%  ($n=653$), with differing mutation frequencies found across different ethnic subpopulations.}
\end{center}
\end{figure*}

Potentially, it may be beneficial to include protected attributes such as ethnicity into algorithms, especially when the target label is inferring genetic variation which is correlated with ancestry. One of the promising deep learning applications in pathology and genomics integration is mutation prediction from Whole Slide Images (WSIs), which, if successful, can be adopted as a low-cost, screening approach for inferring genetic aberrations without high-throughput sequencing\cite{echle2021deep}. A direct clinical application of deep learning-based mutation prediction is to predict biomarkers such as microsatellite instability (MSI), an FDA-approved biomarker for guiding the use immune-checkpoint inhibition therapy, or EGFR in guiding treatment of multiple tyrosine kinase inhibitors (TKI) in lung cancer\cite{kather2019deep}. However, such an approach trained on TCGA and evaluated on the PIONEER cohort may predict low EGFR mutation frequency and misguide Asian patients with incorrect cancer treatment strategies, even with strategies such as importance weighting as the demography size for protected subgroups may be too minuscule. In this particular instance, using protected class information such as ancestry as a conditional label may improve performance on mutation prediction tasks. At the moment, there is no current work on disentangling genetic variation and measuring the contribution of ancestry towards phenotypic variation in the tissue microenvironment, which is precluded by a lack of large, publicly-available, and also multimodal biobank data.

\rjc{The significance of developing diverse biobanks is well-known in other facets of precision medicine research such as genome-wide association studies (GWAS), in which variations in linkage disequilibrium structures and minor allele frequencies across ancestral populations can contribute to worse performance of polygenic risk models in underrepresented populations\cite{duncan2019analysis, lam2019comparative, martin2019clinical, mccarty2011emerge, gottesman2013electronic, manrai2016genetic}. For traits such as schizophrenia disorder, a recent fixed-effect meta-analysis found that despite consistent genetic effects across European (EUR) and East Asian (EAS) populations, polygenic risk models trained on only EUR populations have reduced performance on EAS populations due to differing allele frequencies and associations with the Major Histocompatibility Complex (MHC) signals\cite{lam2019comparative}. Compared to AI algorithms for pathology and other medical imaging domains which have traditionally lacked evaluation on diverse cohorts, trans-ancestry association studies which include populations from divergent ancestries have found enormous success in uncovering new diseased loci that were previously under-powered during detection\cite{duncan2019analysis}. However, an important bottleneck that exists in GWAS studies is still the data interoperability barrier for sharing sensitive health data.}

\vspace{-4mm}
\noindent\textbf{\large{Race-specific Covariate Bias in Risk Calculators}}
\vspace{-4mm}

\rjc{Historical inequities in healthcare is a well-known example of population shift that results in algorithm bias. In the development of risk calculators over the years, spurious associations have been assumed between protected class identity such as race and disease outcome, when in truth, the underlying causal factor stems from social determinants of health and not the class identity itself\cite{burchard2003race, phimister2003race, bonham2018race,vyas2020hidden, obermeyer2019dissecting}. In these cases, racial correlations exist in part because it acts as a proxy for the influence of other socioeconomic factors. When the developers of the eGFR equation noticed a consistent underestimation of kidney function in Black patients, for example, they used a race covariate to adjust for the difference in predictions for Black patients (\textbf{Table 2})\cite{levey1999egfr,levey2006egfr,poggio2005egfr}. Although this covariate improved the prediction accuracy for this cohort on the best available data at the time, the correlational difference may be confounded by a variety of well-known factors such as lack of access to care, delayed screening, or exclusion of certain groups from older research\cite{udler2015effect,eneanya2019egfr}. Furthermore, the race covariate resulted in fewer patients from this cohort being diagnosed with chronic kidney disease and receiving access to specialist referrals, appropriate medications, and transplantation evaluations, which further widens existing health disparities\cite{van2021removing,diao2021removing}. A similar problem is seen in the development of the bone fracture risk assessment calculator, FRAX, for which country-specific and racial corrections are used to account for varying incidence trends of osteoporosis among different populations but may delay intervention with osteoporosis therapy\cite{kanis2018bone,lewiecki2020bone}. In training algorithms on data that have internalized historical biases, algorithms may be learning from these biases and cause disproportionate harm to certain groups of individuals.}

\vspace{-4mm}
\noindent\textbf{\large{Image Acquisition and Measurement Variation}}
\vspace{-4mm}

\begin{figure*}
\includegraphics[width=\textwidth]{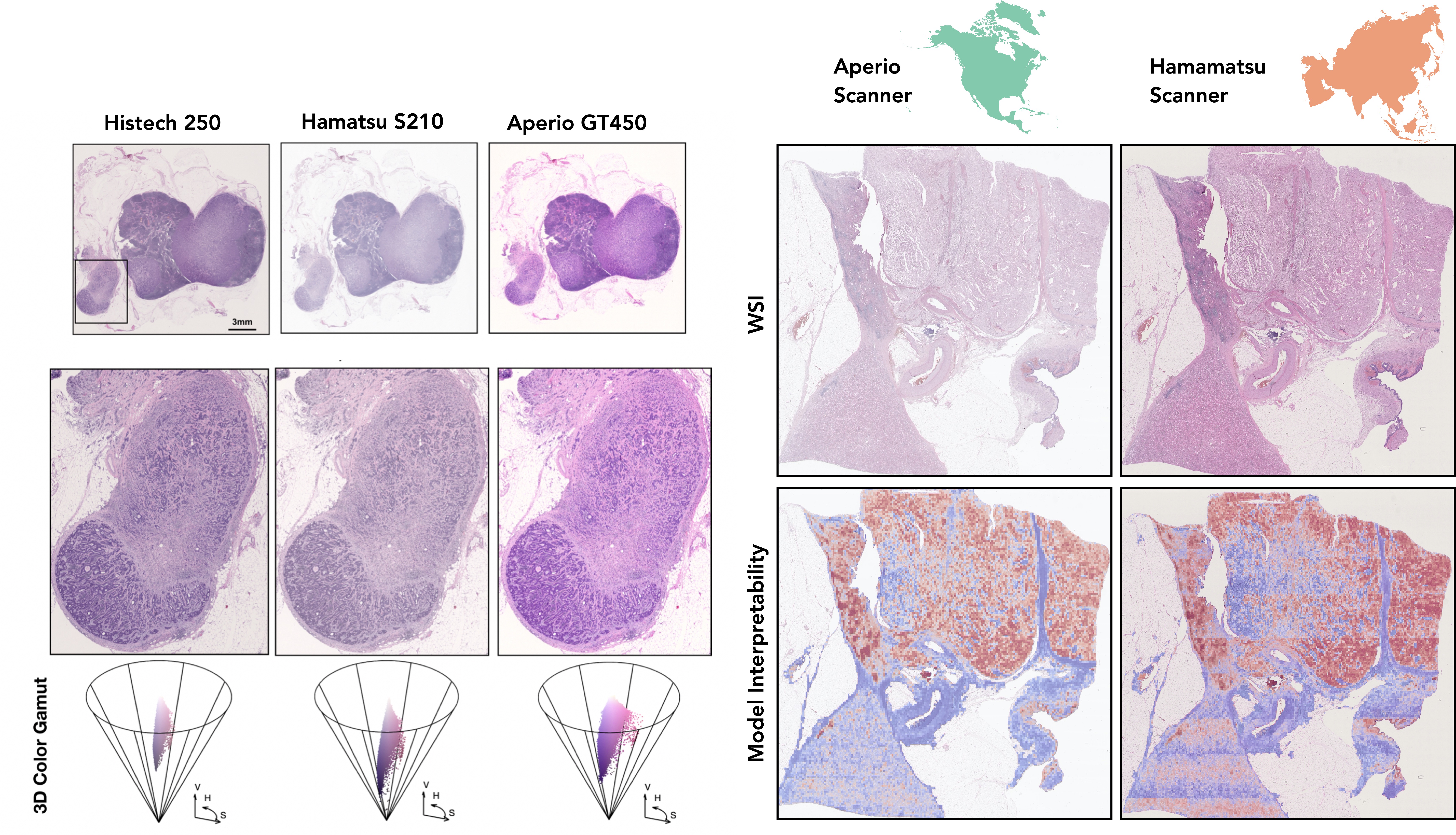}
\caption*{\textbf{Figure 3. Dataset shift in AI-SaMD deployment for a clinical-grade pathology AI algorithm.} Examples of site-specific H\&E stain variability under different whole slide scanners, and their downstream affect on attention-based heatmaps of weakly-supervised AI algorithms in model audit.}
\end{figure*}

In addition to genetic drift in the population, variations in image acquisition and biological measurement techniques can also be confounders that leak protected class information. This type of covariate shift is known as domain (or acquisition) shift, in which patients with the same underlying phenotype and annotation may still vary due to institution-specific protocols and other non-biological factors that affect data acquisition\cite{castro2020causality}. For example, in radiology, collected X-Ray, mammography, or CT imaging data may vary due to radiation dosage which affects the signal-to-noise ratio in producing the image. In pathology, there is also enormous heterogeneity in tissue preparation and staining protocols, as well as as scanner-specific camera parameters for slide digitization, which has been shown to affect model performance in slide-level cancer diagnostic tasks (\textbf{Figure 3}).

Though medical domain shift is a well-recognized problem, domain shift as a result of site/region-specific factors that correlate with demographic characteristics may also introduce spurious associations with ethnicity. For example, a recent audit study assessing site-specific stain variability of pathology slides in the TCGA found shifts in stain intensity in University of Chicago, which notably was the only site with a greater prevalence of patients with African ancestry\cite{howard2021impact}. As a result, many of the aforementioned clinical-grade AI algorithms in pathology may be learning inadvertent cues for ethnicity via institution-specific staining patterns. In this instance of domain shift, variable staining intensity can be corrected using domain adaptation and optimal transport techniques that adapt the test distribution to the training dataset, which can be performed on either the input or representation space. For instance, recent deep learning techniques using generative adversarial networks have been able to learn stain features as a form of style transfer, in which a GAN can be used to preprocess at data at deployment time to match the training distribution\cite{ganin2015unsupervised, shaban2019staingan}. Other in-processing techniques such as adversarial regularization can be leveraged to learn domain-invariant features using semi-supervised learning using samples from both the training and test distributions. However, a practical limitation in both mitigation strategies is that the respective style-transfer or gradient-reversal layers would need to be finetuned with data from the test distribution for each deployment site, which can be challenging to due stagnant data interoperability between institutions as well as regulations for refining AI-SaMDs\cite{us2019proposed}. In some applications, \rjc{understanding} sources of shift presents a challenge in developing bias mitigation strategies that would remove \rjc{unwanted confounding factors}. For instance, despite no known anatomic and phenotype population features in radiology, recent work has found that CNNs can reliably predict race in chest X-ray and radiology images despite controlling for image acquisition factors, removing bone density information and severe degradation of image quality using low- and high-pass filtering\cite{banerjee2021reading}.

\vspace{-4mm}
\noindent\textbf{\large{Evolving Dataset Shifts Over Time}}
\vspace{-4mm}

In medicine, dataset shift can occur also as a result from temporal changes in technology, population and environment, and human behavior, \rjc{which is known as label (or concept) shift}\cite{finlayson2020clinician, guo2021evaluation}. Canonical examples include the migration from ICD-8 to ICD-9 with reclassified the refactored the coding for "surgical" procedure, or the migration from ICD-9 to ICD-10 which created a large spike in opiod-related inpatient stays\cite{tedeschi1984classification, heslin2017trends}. A more recent example of label shift is seen with the Epic Sepsis Model (ESM), a sepsis prediction model that was deactivated in April 2020 due to the changes in patient demographic characteristics confounded by the onset of COVID-19. To mitigate dataset shifts, proposed guidelines have emphasized the importance of guaranteeing model stability to how the data were generated\cite{subbaswamy2020development}, with reactive and proactive approaches for intervening on temporal dataset shift in in active, early-warning systems such as sepsis prediction\cite{guo2021evaluation, subbaswamy2019preventing}.

In the context of fairness, such shifts are difficult to avoid as available data may have been generated through an inherently discriminatory process. Moreover, evaluation of fairness metrics may be difficult at deployment time without access to the test labels, which may further be exacerbated with annotation shift in: 1) intra-observer variability amongst clinicians and 2) evolving clinical knowledge. To date, most work has focused on fine-tuning static pre-trained fair classifiers using few short learning,or developing new fairness measures that address short- and long-term decision-making with multi-task objectives\cite{slack2020fairness, dai2021fair, heidari2019long, wen2021algorithms, heidari2019long, wen2021algorithms}. At the moment, however, analyses of fairness metrics under label shift has not yet been examined in current AI-SaMDs.

\vspace{-4mm}
\noindent\textbf{\large{Fragility of Race}}
\vspace{-4mm}

Similar to the problem of label shift across train and test distributions, different geographic regions and countries may collect protected attribute data with varying levels of \rjc{stringency} and granularity. One issue that complicates the incorporation of race as a covariate in evaluations of fairness of medical AI models is the active evolution of the medical community’s understanding of race itself\cite{board1998aaa}. As discussions regarding race and ethnicity have moved more into the mainstream, the medical community has begun to realize that the racist taxonomies of the past do not adequately represent the groups of people that they purport to. Indeed, it is now accepted that race is a social construct and that there is greater genetic variability within a particular race than there is between races\cite{oni2021embracing, calhoun2021pathophysiology, sun2020don}. As such, categorization of patients by race can obscure a host of potential confounders to fairness analyses including culture, history, and socioeconomic status that all may separately and synergistically influence a particular patient’s health\cite{lannin1998influence, bao2021s}. These manifold factors can also vary by location so that the same person may be considered of different races in different geographic locations, as seen in the example of self-reported Asian ethnicity in the TCGA versus Pioneer and self-reported race in COMPAS\cite{shi2014prospective, bao2021s}.

Ideally, discussions should center explicitly around each component of race and include ancestry, a concept with a clear definition (the geographic origins of one’s ancestors) and one more directly connected to the patient’s underlying genetics. Unfortunately, introducing this granularity to fairness evaluations has clear drawbacks in terms of the power of subgroup analyses and this data is not routinely gathered on patients at most institutions, which often fall back on the traditional dropdown menu that allows one to select only a single race and/or ethnicity. Performing fairness evaluations without explicitly considering these potential confounders of race may mean that the AI system under study is sensitive to some unaccounted-for factor hidden from the analysis\cite{lannin1998influence}.

\noindent\textbf{\Large{Paths Forward}}

\rjc{In examples such as race-specific covariates in race calculators, many conventional fairness techniques developed for tabular data can be applied such as important weighting, data transformation, and variable blinding to correct for confounding features and data curation protocols, as well as the elimination of race-specific covariates as advocated by many experts\cite{diao2021human}. However, fairness techniques cannot be applied as directly to complex structured data modalities such as medical imaging and genomics, which may suffer from many different instances of dataset shift in blackbox AI algorithms. In this section, we highlight several emerging technologies that can be used to develop and evaluate fair AI algorithms and their broader roles in the AI-SaMD lifecycle.}

\noindent\textbf{\large{\rjc{Using Federated Learning to Increase Biobank Diversity}}}


\rjc{Federated learning is a novel distributed learning paradigm in which a network of participating users utilize their own computing resources and local data to collectively train a global model stored on a server\cite{hao2019efficient, yang2019federated, bonawitz2017practical, bonawitz2019towards}. Unlike machine learning performed over a centralized pool of training data, in federated learning, users in principle retain oversight of their own data and instead only have to share the update of weight parameters or gradient signals (with privacy-preserving guarantees) from their locally trained model with the central server. As a result, algorithms can be trained on large and diverse datasets without sharing any sensitive information, and has since been applied to a variety of clinical setting via}: 1) overcoming data interoperability standards that would usually prohibit sensitive health data from being shared: 2) eliminating low data regimes of clinical machine learning tasks that predict rare diseases\cite{brisimi2018federated, huang2019patient, xu2021federated, chakroborty2021beyond, ju2020federated, li2019privacy, kaissis2021end, chen2018my, rieke2020future, sheller2020federated, choudhury2019differential, kushida2012strategies, van2003data, veale2017fairer, fiume2019federated, hernandez2020privacy}. For example, in EMR data, federated learning has been previously demonstrated in satisfying privacy-preserving guarantees for transferring sensitive health data, as well as developing early warning systems for hospitalization, sepsis, and other preventive tasks\cite{choudhury2019differential, duan2018odal}. In radiology, federated learning has been recently used for multi-institutional collaboration and validation of AI algorithms for prostate segmentation, brain cancer detection, and Alzheimer's disease progression monitoring from MRI scans, as well as classification of paediatric chest X-ray classification under various network architectures, privacy-preserving protocols, and ablation studies to adversarial attacks\cite{sarma2021federated, sheller2020federated, li2019privacy, silva2019federated, roy2019braintorrent}. In pathology, model audit studies have assessed the the robust performance of weakly-supervised algorithms for WSIs under various privacy-preserving noise levels in diagnostic and prognostic tasks\cite{lu2020federated}. As a result of the SARS-CoV-2 pandemic in 2019, federated learning has been employed overcoming low sample sizes of COVID-19 pathology in AI model development, as well as in independent test cohort evaluation\cite{dou2021federated, yang2021federated , vaid2021federated}.
\vspace{-4mm}

\rjc{With respect to fairness, federated learning paradigms for decentralized AI-SaMD development can naturally be extended to address many of the aforementioned cases of dataset shift and also mitigate disparate impact via model development on larger and more diverse patient populations\cite{li2021targeting}. For instance, In the case of population shift as a result of genetic variation, decentralized information infrastructure have been previously proposed to harmonize biobank protocols and developed tangible material transfer agreements amongst three hospitals, which demonstrates the potential applicability of federated learning paradigms in developing large and diverse biobank data for diverse populations\cite{mandl2020genomics}. In developing polygenic risk scores, federated learning has been used as an integration strategy in merging heterogeneous population data from multiple healthcare institutions, with subsequent validation of federated models on underrepresented populations\cite{cai2021unified, li2021targeting}. In the previous examples of site-specific staining variability across different hospital sites, federated learning can be used to train decentralized models that are invariant to stain via domain generalization, as well as domain adaptation in refining AI-SaMDs locally to the test data distribution with minimal updates\cite{howard2021impact}. Many of the current research directions in federated learning point towards multi-site domain adaptation across distributed clients, which would naturally mitigate many instances of dataset shift\cite{liang2020we, song2020privacy, li2020multi, peterson2019private, peng2019federated}. For instance, Federated Multi-Target Domain Adaptation (FMTDA) is a recently proposed task that addresses domain gaps between unlabeled, distributed client data and labeled, centralized data over the server, as well as degraded performances of federated domain adaptation methods\cite{yao2021federated}. In application to other areas of fairness, opportunities created via federated learning may allow for novel data preprocessing strategies such as importance weighting in reweighting not only infrequent samples, but also model updates from clients containing only under-represented minority subpopulations. In addition, fairness evaluation criteria can potentially be evaluated at the client-level, which may contribute towards developing other novel in-processing and post-processing techniques without knowledge of protected attributes. In instances such as multi-site TCGA data which leak ethnicity subgroup membership largely due to the correlation of sample selection bias and H\&E stain variability, many fairness evaluation criteria and post-processing techniques can potentially be performed at the client-level without detailed knowledge of patient demography\cite{howard2021impact}}.

\noindent\textbf{\large{\rjc{Operationalizing Fairness Principles across the Healthcare Ecosystem}}}

Though federated learning may overcome data interoperability standards and enable training AI-SAMDs with diverse cohorts, the evaluation of AI biases in federated learning settings is yet to be extensively studied. Despite numerous technical advances made in improving communication efficiency, robustness and security of parameter updates, one of the key statistical challenges is learning from non-i.i.d data, which arises due to the sometimes vast differences in local data distribution at contributing sites, which can lead to the divergence of local model weights during training following a synchronized initiation of model weights\cite{zhao2018federated, bonawitz2017practical, konevcny2016federated, lin2017deep}. Accordingly, the performance of FL algorithms, including the well-known FedAvg algorithm\cite{mcmahan2017communication, li2018federated} that uses averaging to aggregate local model parameter updates has been shown to deteriorate substantially when applied to non-i.i.d. data\cite{sattler2019robust}. Such statistical challenges may produce further disparate impact depending on heterogeneity of data distributions across clients. For instance, in using multi-site data in the TCGA as individual clients for federated learning, for the TCGA Invasive Breast Carcinoma (BRCA) cohort, a majority of parameter updates would come from clients that over-represent individuals with European ancestry, with only one parameter update coming from a single client that has majority representation for African ancestry. For decentralized training with diverse biomedical datasets, an important consideration is thus the added complexity in disentangling the impact of site-specific dataset shift from non-i.i.d. data on algorithm fairness, which will not only be problem-specific but also highly-variable from the collaboration structure of participating institutions from differing geographic locations. Moreover, federated learning models can still be affected by biases in the local dataset of each participating client or institution, as well as other variables such as the weighted contribution of each site in updating the global model and the varying frequencies at which different sites participate in training\cite{abay2020mitigating}. In the previous example of multi-site data from the TCGA-BRCA cohort, federated models would be subject to site-specific biases such as H\&E stain intensity, intra-observer variability, ethnic minority under-representation found in centralized models\cite{howard2021impact}. Though federated learning would enable the application of finetuning AI-SaMDs per deployment site, race/ethnicity-evaluation of federated models has yet to be benchmarked. Lastly, access to protected attribute data in each site may still pose in issue in fairness evaluation, as sensitive information such as race and ethnicity are typically isolated in separate databases and may produce additional logistic barriers despite interoperability.

\rjc{A final but important consideration in the practical adoption of fair and federated learning paradigms is the current difficulty for practitioners in operationalizing fairness principles for much simpler AI development and deployment life-cycles of centralized models. In current organizational structures, the roles and responsibilities created for implementing fairness principles are typically isolated into: 1) practitioner / "data regulator" roles, which design AI fairness checklists for guiding the ethical development of algorithms in the organization, and 2) engineer / "data user" roles, which follow the checklist during algorithm implementation\cite{mcnamara2019costs}.} Though intuitive, such binary roles may have poor efficacy in practice, as fairness checklists are often too broad, abstractive, and not co-designed with engineers in addressing problem-specific, technical challenges for achieving fairness\cite{madaio2020co}. \rjc{In the consideration of federated learning paradigms for AI-SaMD development and deployment at a global-scale, the design of fairness checklists would require not only interdisciplinary collaboration from all healthcare-related roles (\textit{e.g.} clinicians, ethics practitioners, engineers, researchers), but also further involvement from stakeholders at participating institutions in identifying potential site-specific biases that may be propagated during parameter sharing or accuracy-fairness tradeoffs at inference time\cite{mcnamara2019costs}. Overall, the potential of federated learning presents a new opportunity to evaluate AI algorithms on diverse biomedical data at a global-scale, but faces unknown challenges in the design of global fairness checklists that would understand the burdens and patient preferences of each region. For instance, in a hypothetical setting in which a federated scheduling algorithm for patient follow-ups is calibrated to set a high threshold to maximize fairness criteria, a low-resource setting may be much more overburdened with patient follow-ups than that a high-resource setting\cite{jung2021framework}. Similar to the problem of label / annotation shift that may occur at various sites, there would exist additional complexity in considering culture-specific factors within each geographic region that would affect access to protected information, as well as definitions and criteria for fairness from differing moral and ethical philosophies\cite{awad2018moral}. Though challenging at an operational level, incorporating preferences from diverse stakeholders (and in particular, underrepresented populations) are ultimately needed to navigate such ethical conflicts.}

\begin{figure*}
\vspace{-9mm}
\includegraphics[width=\textwidth]{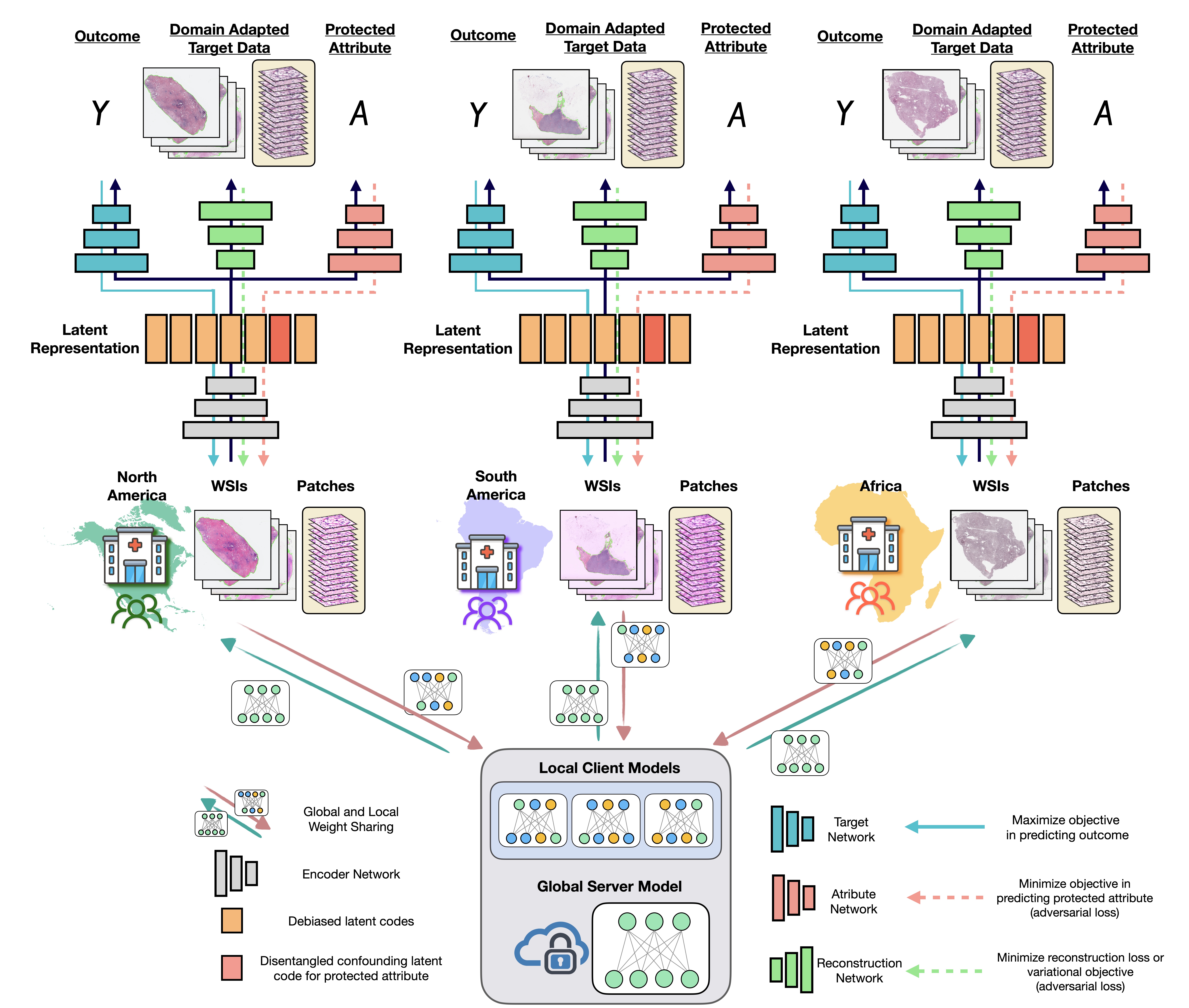}
\caption*{\textbf{Figure 4. A decentralized framework that integrates federated learning with adversarial learning and disentanglement.} In addition to developing algorithms on larger and more diverse patient populations, federated learning can also be integrated with many current techniques in representation learning and unsupervised domain adaptation that can additionally learn with unobserved protected attributes. In FADE, the client IDs are used as protected attribute, with adversarial learning used to debias the representation to be invariant to geographic region (attribute network branch, colored red)\cite{hong2021federated}. In FedDis, disentanglement was used to disentangle shape and apperance features in brain MRI scans, with only the shape parameter shared between clients (disentangled representation, colored orange)\cite{bercea2021feddis}. In FADA, disentanglement and adversarial learning can be used to further mitigate domain shift across clients (both attribute network branch and disentangled representation, colored red and orange)\cite{ke2021style}. Federated learning can also be used in combination with style transfer, synthetic data generation, and image normalization in which domain adapted target data would need to be shared (reconstruction network branch, colored green)\cite{ke2021style, pfohl2019federated, xin2020private, rajotte2021reducing}.}
\end{figure*}

\noindent\textbf{\large{\rjc{Fair Representation Learning}}}

\rjc{Fair representation learning is a new direction in deep learning which focuses on learning intermediate representations that retain discriminative features from the input space $X$ without any features that correlates with $A$ via an adversarial loss term, as seen in \textbf{Figure 1}. Fair representation learning is also orthogonal to many related fields such as causality and model robustness, sharing similar techniques in using adversarial learning for debiasing representations and having similar goals in eliminating harm for downstream tasks. Inspired by the minimax objective in Generative Adversarial Networks (GANs), adversarial learning has since been used in dataset shift literature in learning domain-invariant features of the train and test distribution, \textit{e.g. -} $P_{\text{train}}(Y|X) = P_{\text{test}}(Y|X)$\cite{ganin2015unsupervised}, in causality to learn treatment-invariant representations for producing counterfactual explanations\cite{louizos2017causal, shi2019adapting, luo2020causal, scholkopf2021toward}, and in fairness in learning race/ethnicity-invariant features to removing disparate impact of deep learning models\cite{celis2019improved, zhao2019inherent, rezaei2020fairness, petrovic2020fair, sattigeri2019fairness, xu2018fairgan, xu2021robust, wadsworth2018achieving, adel2019one}.}

\rjc{Though fair representation methods are typically supervised, training fair AI algorithms in an unsupervised manner would allow representations to be freely transferred to other domains without constraints on downstream classifiers as being fair, enabling greater applications of fair algorithms without protected access to attributes\cite{zemel2013learning, madras2018learning}. One prominent example is the "Learned Adversarially Fair and Transferable Representation" (LAFTR) method, which was the first to propose modifying the GAN minimax objective with an adversarial loss term to make the latent feature representation invariant to protected class. Moreover, LAFTR also showed that such representations are transferable, as examined in the Charlson Comorbidity Index prediction task in the Heritage Health dataset, in which LAFTR able to transfer to other tasks without leaking sensitive attributes\cite{zemel2013learning, madras2018learning}. Across other tasks in medicine, the novelty of LAFTR can be extended as a privacy-preserving machine learning approach that allows the transfer of useful intermediate features, which could advance multi-institutional collaboration in fine-tuning algorithms without leaking sensitive information. Similar to LAFTR are methods in unsupervised fair clustering, which further evaluates the fairness of debiased representations (via adversarial learning) in achieving attribute-invariant cluster assignments. Still, a key limitation in many of these unsupervised fairness approaches is that they depend on having the protected attribute at-hand during training, which may not be possible in many clinical settings in which protected class identity such as ethnicity is secured in many healthcare systems. Moreover, in assessing the accuracy-fairness trade-off, adding additional regularization components may decrease representation quality and thus lower performance in downstream fairness tasks\cite{pfohl2021recommendations}.}

Though access to protected attributes may pose in issue in model training, a potential advantage that may benefit the design of fair and also federated models is that the geography of the client identities may be used as proxy variables for subgroup identity, which may inform the development of novel fairness techniques without access to sensitive information. Recent decentralized frameworks have demonstrated that federated learning, in combination with fair representation learning, can be used to learn federated, adversarial, debiasing (FADE) representations with similar privacy-preserving and transferable properties as LAFTR, which was benchmarked on mild cognitive impairment detection from sensor data\cite{hong2021federated}. In comparison to other paradigms, FADE as a unique property in that protected attributes are not needed to learn invariant representations, as client identities can be used instead during adversarial regularization. Overall, in training with heterogeneous data sources, such an assumption may hold for many clinical tasks as geography has been shown to be a closer link to genetic diversity than ethnicity\cite{manica2005geography}. 

\noindent\textbf{\large{{\rjc{Debiased Representations via Disentanglement}}}}

Tangential to the work of unsupervised fair representation learning in LAFTR and FADE is disentanglement in generative models, which can also be used to further promote fairness in learned representations without access to protected attributes. Disentanglement is a growing research subfield within representation learning which aims as disentangling independent and easy-to-interpret factors of data in the latent space, and has demonstrated success in isolating sources of variation of objects such as color, pose, position, and shape\cite{hadad2018two, achille2018emergence, chen2018isolating, kim2018disentangling, higgins2016beta}. The first method to demonstrate and quantify disentanglement in deep generative models was BetaVAE, which uses a variational autoencoder (VAE) bottleneck for unsupervised learning, followed by proposing a disentanglement score via training a linear classifier to predict the fixed factor of variation from the representation\cite{higgins2016beta}. \rjc{In application to medicine, disentangled VAEs with adversarial loss components have been used in disentangling size, skin color, and eccentricity in dermoscopy images, as well as causal health conditions and anatomical factors in physiological waveforms\cite{sarhan2019learning, gyawali2021learning, bing2021disentanglement}.} 

In relation to fairness and dataset shift, disentanglement can be viewed as a form of data preprocessing that can not only de-bias representations for downstream fairness tasks, but also provide flexibility in allowing the data user to isolate and truncate specific latent codes that correspond to protected attributes in the representation space\cite{cisse2019fairness, mcnamara2019costs, creager2019flexibly}. Recent exhaustive studies on the evaluation of unsupervised VAE-based disentangled models have demonstrated that disentanglement scores have correlated with fairness metrics, benchmarked on numerous fair classification tasks without protected attribute information\cite{locatello2019fairness}. \rjc{In application, disentanglement would be particularly advantageous in settings with many protected attributes and different downstream tasks, such as in pathology where ethnicity may be excluded from slide-level features in predicting cancer stage but included in predicting mutation status\cite{creager2019flexibly}.} In human face data, disentanglement-like methods have been proposed in clustering faces without latent code information that contain dominant features such as as skin and hair color\cite{li2020deep}. In application to federated learning and medical imaging, frameworks such as FedDis has been demonstrated to isolate sensitive attributes in non-i.i.d. MRI lesion data, in which images are disentangled into shape and appearance features with only the shape parameter shared between clients (\textbf{Figure 4})\cite{bercea2021feddis}. 

%
 




\noindent\textbf{\large{\rjc{Disentangling the role of ``Data Regulators" and ``Data Users" in AI-SaMD Lifecycles}}}

\rjc{Within current development and deployment lifecycles for AI-SaMDs and other AI algorithms, the adaptability of unsupervised fair representation and disentanglement methods can potentially be used in refining the distribution of responsibilities in organizational structures in also including the role of ``data producers", who produce ``cleaned" versions of the input that are still informative in downstream tasks\cite{mcnamara2019costs}. In this setting, the role of ``data users" would be separated from that of ``data regulators", which may allow conventional model development pipelines without considering additional fairness constraints. Moreover, the role of the ``data producers" would be in also quantifying the potential accuracy-fairness trade-offs in using regularization components for achieving debiased and disentangled representations. Such an approach has been hypothesized to pave a path forwards for a novel three-party governance model that simplifies communication overhead in discussing concerns of accuracy-fairness trade-offs, and may adapt to test populations without the complexities introduced by federated learning which also needs access to protected attributes. Overall, though fair representation learning offers potential paths forward in making classifiers more flexible without needing protected attributes, current methods have yet to be benchmarked against competitive self-supervised learning methods (\textit{e.g.} - contrastive learning), with more evaluation in clinical settings needed\cite{chen2020simple}. As an ongoing and promising research direction, future work into understanding disentanglement and adapting it robust self-supervised learning paradigms would contribute to improving fairness in transfer learning tasks, as well as also serving as a privacy-preserving measure that would be useful in clinical machine learning tasks.} 

\noindent\textbf{\Large{Algorithm Interpretability for building Fair and Trustworthy AI}}

\rjc{In current regulatory frameworks for AI-SaMD development, interpretability serves a pivotal role in not only interpretation of AI algorithms in medical decision-making, but also model auditing in understanding sources of unfairness and detecting dataset shift\cite{shad2021designing}. Tangent to interpretability and fairness is the notion of trust in AI algorithms, which is an important criteria in current regulation of AI-SaMDs, Trust is a recent conceptualization in the machine learning community and now broadly advocated by regulatory bodies such as the European Commission and the FDA, in which "trust" is the fulfillment of a contract in human-AI collaboration, and "contracts" are AI functionalities that are anticipated by the human with known vulnerabilities\cite{jacovi2021formalizing}. For instance, model correctness is a contract that anticipates "patterns that distinguish the model’s correct and incorrect cases are available to the user", with many different types of contracts (related to technical robustness, safety, non-discrimination, transparency) outlined by the European Commission on ethical guidelines for trustworthy AI\cite{floridi2019establishing,hleg2019ethics}. Similarly, the FDA have published action plans for developing trust in AI-SaMDs, and have specifically identified bias assessment and interpretability as contracts within recent guidelines on "Good Machine Learning Practices"\cite{us2021artificial}. In the subsections below, we briefly review interpretability methods, and their current discussion in fairness and medicine.}

\noindent\textbf{\large{Interpretability Methods and Model Auditing Usages}}




\noindent\rjc{\textit{Imaging data}: In imaging data, class activation maps (CAM or saliency mapping) are commonly used to find sensitive input features that would explain the decision made by a network. Similar to Integrated Gradients, these methods compute the partial derivatives of the predictions with respect to pixel intensities computed during back-propagation of the network, which are then able to produce a visualization of informative pixel regions\cite{simonyan2013deep}. To produce more fine-grained visualizations, extensions such as Grad-CAM instead attribute how neurons of an intermediate feature layer of a CNN would affect the output, such that the attributions for these intermediate features can be  upsampled to the original image size and viewed as a mask to identify discriminative image regions\cite{selvaraju2017grad}. Within the medical imaging community, CAM-based methods have gained widespread adoption in the interpretation of CNNs for clinical interpretability, as salient regions would refer to high-level image features rather than low-level pixel intensities. Because these techniques can be applied without modifying the networks, existing and/or deployed models can be readily adapted to visualize network predictions, and have since been used to interpret models in various AI-based medicine applications such as skin lesion detection, chest x-ray disease localization, and CT organ segmentation\cite{sayres2019using, patro2019u, irvin2019chexpert, grewal2018radnet, arun2020assessing, poplin2018prediction, pierson2021algorithmic, schlemper2018attention, schlemper2019attention, oktay2018attention}.}.

\rjc{Despite their popularity in the medical imaging and broader vision communities, saliency mapping techniques have also received longstanding scrutiny as their interpretability may not be accurate, human-understandable, actionable, and thus trustworthy by practitioners in medical support decision-making, biomarker discovery, and model auditing applications\cite{mittelstadt2019explaining,kindermans2019reliability,kaur2020interpreting}. In other words, saliency mapping for detecting meaningful visual concepts is qualitatively interpretable but not quantifiable in using these explanations to evaluate group differences. In natural images as well as echocardiograms, recent audit studies of Grad-CAM interpretability have found saliency maps to be misleading and often equivalent to results from simple edge detectors\cite{shad2021designing, adebayo2018sanity}. In chest radiograph diagnosis, a wide variety of saliency mapping techniques have been demonstrated to have poor AUC performance in localizing pathologic features\cite{arun2020assessing, saporta2021benchmarking}. In using interpretability for biomarker discovery, whereas applying Integrated Gradients to genomics data can readily detect feature importance of IDH1 mutation in diffuse gliomas (and demonstrating prognostic value via stratifying IDH-wild-type and IDH-mutant patients into separate survival distributions), it is less clear how highly-attributed pixel regions can be similarly used for patient stratification without further post-hoc assessment\cite{chen2020pathomic}. Lastly, in the previous problem of unknown dataset shift of radiology images leaking self-reported race, model auditing via saliency mapping was highlighted to be ineffective in pointing towards explainable anatomic landmarks or image acquisition factors in causing misdiagnosis\cite{banerjee2021reading}. Though not always informative for medical interpretation, recent works have demonstrated that saliency mapping can additionally be used to detect spurious bugs / artifacts in the input feature space which would create "shortcuts" in AI algorithms, but still being limited in detecting contaminated models\cite{degrave2021ai, adebayo2020debugging}. In highlighting these examples, we conclude in noting that though the visual appeal of saliency mapping may be informative for some medical interpretation usages, its many ambiguities prevents trust development with AI algorithms and ethical adoption in evaluating fairness for clinical deployment scenarios\cite{jacovi2021formalizing, lee2021included}.}

\noindent\rjc{\textit{Tabular Data}: In tabular-structured data, techniques such as Shapley Additive Explanations (SHAP) and Integrated Gradients have gained ubiquity in explaining machine learning predictions across a variety of applications\cite{sundararajan2017axiomatic,lundberg2017unified,ghassemi2021false}. In genomics data, the use of Integrated Gradients has been demonstrated to corroborate the role of impact cancer genes in cancer survival and influential single nucleotide polymorphisms in GWAS\cite{chen2020pathomic, sharma2020deep}. Importantly, these techniques attribute features at both a global-level across the entire dataset (for assessing overall feature importance) and a local-level for individual samples (for explaining individual predictions), and thus can be intuitively extended to model auditing of global and individual fairness criteria respectively\cite{wexler2020probing}. For instance, in SHAP, feature-level importances are computed via decomposing the model output into set of attribution units, in which each attribution relates to the influence of its respective feature on the model output, with the summation of the attributions producing the model output. As a result, group fairness metrics (computed as performance differences of model outputs across protected subgroups) can similarly be decomposed in quantifying the impact of influential features on disparity measures\cite{lundberg2020explaining}. On the MIMIC dataset, a recent analysis examining mortality prediction through the dual lens of interpretability and fairness found disparities in feature importance across ethnicity, gender and age subgroups, with ethnicity often ranked as one of the most important features across a diverse set of models and explainability techniques\cite{meng2021mimic}. On nutrition data from the National Health and Nutrition Examination Survey (NHANES), a causal variation of SHAP was able to discover causal relationships between race and access to food programs in predicting 15-year survival rates\cite{pan2021explaining}. In other model auditing applications, a recent analysis found that not only do greater disparity measures correlate with larger SHAP values for the biased features, but following bias mitigation via importance weighting, also lower disparity measures correlate with lower SHAP values\cite{cesaro2019measuring}}.


\noindent\textbf{\large{Fitting Algorithm Design into AI-SaMD Explainability Compliance}}

\rjc{Though algorithm interpretability techniques are model-agnostic, the efficacy and usages of these techniques vary across different types of data modalities and model architectures, and thus have important implications in the regulatory assessment of human-AI trust for AI-SaMDs and choice of algorithm design. For instance, structured modalities such as imaging data are can be difficult to interpret and trust by clinical and ML practitioners, as feature attributions computed for influential pixels in a CNN are not meaningful in explaining quantifiable disparity measures. As such, regulatory agencies that would enforce specific contracts for building trustworthy AI (\textit{e.g.} - quantifying feature importance for fairness disparity measures) may potentially constrain the design of algorithms that would see clinical deployment, potentially moving away from certain types of deep learning approaches that have limited interpretability. In fact, an interesting trend in recent works have highlighted using human-interpretable and handcrafted features in machine learning models rather than deep learning approaches. For instances, in introspecting digital biomarkers for mild cognitive impairment using mobile devices, a recent study proposed extracting handcrafted minute-, hour-, and day-level statistics from sensor streams as input into a XGBoost classifier followed by SHAP interpretability\cite{chen2019developing}. In pathology, statistical contour- and image-based cell features were successful in predicting molecular signatures such as immune checkpoint protein expression and homologous recombination deficiency\cite{diao2021human}. As an approach for mitigating and explaining unfairness, handcrafted features for predicting prostate cancer recurrence have found success in not only corroborating stromal morphological with aggressive cancer phenotypes, but also elucidating a potential novel, population-specific phenotype for African American patients\cite{bhargava2020computationally}. Though using simpler methods, the above examples highlight important consideration in which handcrafted features were chosen over deep learning-based approaches due to having more informative clinical interpretability and greater trust in model explanations. Overall, as artificial intelligence continues to cross the precipice into clinical workflows, using both deep and handcrafted features alike, the assessment of harm using interpretability will play fundamental roles in refining FDA regulatory frameworks for AI-SaMDs.}

\end{spacing}

\begin{nolinenumbers}
\vspace{-5mm}

\section*{References} 
\vspace{2mm}

\begin{spacing}{1.1}
\bibliographystyle{naturemag}
\bibliography{main_paper.bib}
\end{spacing}
\end{nolinenumbers}
\newpage
\noindent\textbf{\Large{Supplementary Material}s}
\begin{spacing}{1.45}
\noindent\textbf{\large{Overview of Fairness Techniques}}

To reduce violation of group fairness, many techniques have been developed that adapt existing algorithms with: 1) pre-processing steps that remove, augment, or reweight the input space to eliminate confounding bias\cite{celis2019improved, kamiran2012data, krasanakis2018adaptive, jiang2020identifying, chai2016unsupervised}, 2) in-processing steps that construct additional optimization constraint or regularization loss functions that penalize non-discrimination\cite{kamishima2012fairness, zafar2017fairness, goel2018non, agarwal2018reductions}, and 3) post-processing steps that apply correction to calibrate model predictions across subgroups\cite{kleinberg2016inherent, corbett2018measure, pleiss2017fairness, chouldechova2018case}. We briefly review several below:
\subsubsection*{Preprocessing}
\vspace{-4mm}
Algorithmic biases in healthcare stem from historical inequities that create spurious associations would link protected class identity to disease outcome in the dataset, when the true underlying causal factor stems from poor social determinants of health. In training algorithms that on health data that have internalized such biases, the distribution of outcomes across ethnicity may be skewed in which under-served Hispanic and Black patients have more delayed referrals for cancer screening, which may result in more high-grade, invasive phenotypes at the time of radiology imaging or tumor biopsy. Such sources of labeling prejudice are known as negative legacy, or sample selection bias, in which biased data curation protocols may induce correlations between protected attributes and other features, and may result in failed convergence of the training algorithm\cite{kamishima2012fairness}. As a result, many data preprocessing steps have been developed beyond "fairness through unawareness", such as importance weighting, resampling, data transformation, and variable blinding that would correct for confounding features and data curation protocols.

\noindent\textit{Importance Weighting:} Importance weighting first emerged as an approach for eliminating covariate shift, \textit{e.g. - } $P_{\text{train}}(X) \neq P_{\text{test}}(X)$, in which the train distribution is reweighted to match the test distribution via computing a density ratio $\frac{P_{\text{train}}(X)}{P_{\text{test}}(X)}$\cite{sugiyama2007covariate, buda2018systematic}. In fairness, importance weighting is used to reweight infrequent samples belonging to protected subgroups followed by optimization of fairness metrics such as TPR and FPR for that subgroups\cite{celis2019improved, kamiran2012data, krasanakis2018adaptive, jiang2020identifying, chai2016unsupervised}. For clinical tasks in which the distributions for demography or disease prevalence may not match in the train and test population, importance weighting has been extensively used to correct for sample selection bias. For instance, In MRI scans, importance weighting has been previously applied to reweigh instances of sparsely annotated voxels in Alzheimer's disease diagnosis\cite{goetz2015dalsa, wachinger2016domain}. In skin lesion classification, reweighting approaches have similarly been used for learning with noisy labels\cite{xue2019robust}. A noticeable limitation is that classifiers trained with reweighted samples may not have robust performance on multiple domains, as well as can induce high variance in the estimator for severely underrepresented subgroups\cite{globerson2006nightmare, cortes2010learning}.

\noindent\textit{Targeted Data Collection and Resampling:} In practice, many empirical and real-world studies have found that increasing the size of the dataset is able to mitigate biases\cite{chen2018my}. Though audits of current publicly-available and commercial AI algorithms have revealed large performance disparities, collecting data for the under-represented subgroup is able to reduce performance gaps\cite{abernethy2020adaptive, raji2019actionable}. However, such targeted data collection may pose ethical and privacy concerns as a result of additional surveillance, as well as practical limitations especially in collecting protected health information due to stringent data interoperability standards\cite{rolf2021representation}. Similar to importance weighting, resampling aims to correct for sample selection bias via obtaining more fair subsamples of the original training dataset, and can be intuitively applied to correct for under-representation of subgroups\cite{abernethy2020adaptive, iosifidis2018dealing, vodrahalli2018all, barocas2016big, o2016weapons}. However, a computational challenge in fair resampling is maintaining feature diversity, in which over-sampling to correct for under-representation may instead decrease feature diversity. As a result, the development of frameworks for understanding data subsampling has emerged as its own subfield tangent to fair machine learning. One such framework is determinantal point process ($k$-DPP), which proposes quantifiable measures for combinatorial subgroup diversity (via Shannon Entropy) and geometric feature diversity (via measuring the volume of the $k$-dimensional feature space)\cite{celis2016fair}. In tandem with resampling is the problem of optimal data (or resource) allocation in operationalizing dataset collection in statistical surveys, as well as game-theoretic frameworks for understanding the influence of individual data points via Shapley values, which may refine resampling techniques to developing fair and diverse training datasets\cite{lohr2009sampling, chawla2002smote, pukelsheim2006optimal, cesaro2019measuring}. For instance, group distributionally robust optimization (GDRO) is a technique developed to minimize the maximum empirical risk over subgroups with respect to fairness definitions such as disparate impact and mistreatment minimization, and has recently been shown to adapt well to medical imaging tasks such as skin lesion classification\cite{hu2018does, sagawa2019distributionally, rolf2021representation}. Though data allocation would not fix datasets with labeling prejudice, allocations could be used to audit the training set used to develop algorithms for biases.

\vspace{-4mm}
\subsubsection*{In-processing}
\vspace{-4mm}
In addition to eliminating disparate impact of algorithms from the input space, the bulk of many approaches adopt a regularization or adversarial loss term within the model that penalize learning discriminatory features in $X$ in predicting outcomes. For example, in structured data modalities such as imaging, many medical imaging modalities such as radiology, pathology, and even fundus photography images have been showed to leak and detect age, gender, and race from subtle cues in the input space\cite{poplin2018prediction, howard2021impact, banerjee2021reading}. These methods can be separated into two classes: 1) constraint optimization approaches that directly impose a non-discrimination term on the learning objective of a probabilistic discriminative model, and 2) fair representation methods, which are generally deep learning-based and can be unsupervised unsupervised in learning invariance to $A$. An overview of fair representation learning is presented in the \textbf{Paths Forward section}.

\noindent\textit{Constraint Optimization.} As suggested, constraint optimization approaches typically include a non-discrimination term in the learning objective of models such as logistic regression classifiers and support vector machines to fulfill disparate impact and treatment mitigation\cite{kamishima2012fairness, kamishima2011fairness1, zafar2017fairness, zafar2017fairness2, goel2018non}. In logistic regression models, loss terms can be created via computing the covariance of the protected attributes with the signed distance of the sample's feature vectors $X$ to the decision boundary, or modifying the decision boundary parameters to maximize fairness (minimizing disparate impact or mistreatment) subject to accuracy constraints\cite{zafar2017fairness}. Modifications to stochastic gradient descent have also been proposed to weigh fairness constraints in online learning as well\cite{kim2018fairness}. A large limitation in constraint optimization approaches is that the learning objective is made non-convex in including additional non-discriminatory terms, which have been shown to reduce performance in comparison to simple reprocessing techniques such as resampling and importance weighting.

\vspace{-4mm}
\subsubsection*{Post-processing}
\vspace{-4mm}
Calibration is a post-processing technique used to satisfy sufficiency in fairness, and is applied to equalize the proportion of positive predictions to that of true positives in classification problems (Predictive Quality Parity). As demonstrated in \textbf{Box 2}, calibration is applied across all protected subgroups to ensure that probability estimates carry the same meaning across subgroups. As a risk tool, calibration theoretically allows risk estimates to have the same independent effectiveness regardless of group membership as seen in in COMPAS. Despite this desirable role in risk assessment, calibration has been extensively studied and demonstrated to not always imply non-discrimination\cite{kleinberg2016inherent}. For instance, redlining in banking is an example of a calibrated algorithm that strategically misclassifies individuals with the intention of discrimination\cite{corbett2018measure}. Calibration has also been shown to be incompatible with alternative definitions of fairness such as equalized odds and disparate impact outside of highly constrained cases\cite{pleiss2017fairness, chouldechova2018case}. To reconcile error parity and calibration, withholding predictive data for randomized inputs is often used as a post-processing step\cite{pleiss2017fairness, liu2019implicit}. However, the exclusion of individual predictions and trade-off in model accuracy highly disfavor the use of this method in criminal justice and healthcare systems. Growing literature has begun exploring complementary concepts in fairness such as multi-calibration, an approach that focuses on identifiable subpopulations of individuals rather that large sets of protected groups. By using small samples of training data, multi-calibration situations where predictions at an individual level are considered the most fair\cite{kearns2019empirical, hebert2018multicalibration}. 
\end{spacing}

\end{document}